\documentclass[runningheads]{llncs}

\usepackage{eccv}

\usepackage{eccvabbrv}

\usepackage{graphicx}
\usepackage{booktabs}
\usepackage{multirow}
\usepackage{pifont}
\usepackage{caption}

\usepackage{xcolor, colortbl}
\definecolor{LightCyan}{rgb}{0.88,1,1}
\definecolor{Gray}{gray}{0.4}
\definecolor{LightCyan}{rgb}{0.82,0.82,1}
\definecolor{LightGray}{HTML}{E3E4E4}
\definecolor{Pearl}{HTML}{F1EAE3}
\definecolor{yellow-small}{HTML}{FCF3CF}  
\definecolor{purple-med}{HTML}{E8DAEF}
\definecolor{green-large}{HTML}{A9DFBF}
\definecolor{tableblue}{HTML}{98AFC7}

\usepackage[accsupp]{axessibility}  

\usepackage{hyperref}

\usepackage{orcidlink}

\begin{document}

\title{Track Any Peppers: Weakly Supervised Sweet Pepper Tracking Using VLMs}

\titlerunning{Track Any Peppers}

\author{Jia Syuen Lim \orcidlink{0009-0008-0003-4805} \and
Yadan Luo  \orcidlink{0000-0001-6272-2971} \and
Zhi Chen \orcidlink{0000-0002-9385-144X} \and
Tianqi Wei \orcidlink{0009-0005-0134-6438} \and \\
Scott Chapman \orcidlink{0000-0003-4732-8452} \and
Zi Huang \orcidlink{0000-0002-9738-4949}}

\authorrunning{J. Lim et al.}

\institute{The University of Queensland, Australia \\ 
\email{\{jiasyuen.lim, y.luo, zhi.chen, tianqi.wei, scott.chapman, helen.huang}@uq.edu.au\}}

\maketitle

\begin{abstract}
  In the \textit{Detection and Multi-Object Tracking of Sweet Peppers Challenge}, we present \underline{T}rack \underline{A}ny \underline{P}eppers (\textbf{TAP}) - a weakly supervised ensemble technique for sweet peppers tracking. TAP leverages the zero-shot detection capabilities of vision-language foundation models like Grounding DINO to automatically generate pseudo-labels for sweet peppers in video sequences with minimal human intervention. These pseudo-labels, refined when necessary, are used to train a YOLOv8 segmentation network. To enhance detection accuracy under challenging conditions, we incorporate pre-processing techniques such as relighting adjustments and apply depth-based filtering during post-inference. For object tracking, we integrate the Matching by Segment Anything (MASA) adapter with the BoT-SORT algorithm. Our approach achieves a HOTA score of 80.4\%, MOTA of 66.1\%, Recall of 74.0\%, and Precision of 90.7\%, demonstrating effective tracking of sweet peppers without extensive manual effort. This work highlights the potential of foundation models for efficient and accurate object detection and tracking in agricultural settings.
\end{abstract}

\section{Introduction}
\label{sec:intro}
The integration of computer vision technologies into horticulture has become increasingly vital for modern agricultural practices. Accurate detection and tracking of small objects, such as sweet peppers in densely populated fields, is essential for monitoring crop health, identifying diseases \cite{wei2024benchmarking, wei2024snap, wei2024plantseg}, assessing harvest readiness, phenotyping \cite{Chen2024CFPRNet} and making informed decisions that enhance sustainability and production efficiency \cite{halstead2021crop, smitt2024pagnerf, smitt2021pathobot}. Automating these tasks not only reduces the reliance on manual labor but also improves precision and scalability in crop management.

Traditional object tracking algorithms require extensive training on large, annotated datasets \cite{masa}. This process involves manually assigning instance tracking IDs across video frames in addition to labeling bounding boxes—a labor-intensive and time-consuming endeavor that can be prohibitively expensive. The need for such detailed annotations makes it impractical to frequently update models to adapt to the dynamic conditions commonly found in agricultural environments.

Recent advancements in large foundation models have showcased remarkable zero-shot and generalization capabilities, particularly in vision-language models (VLMs) like Grounding DINO \cite{liu2023grounding} and Segment Anything Model (SAM) \cite{kirillov2023segment}. These models can perform object detection without task-specific training data, presenting an opportunity to mitigate the extensive manual effort typically required for dataset annotation.

In this work, we propose a novel methodology that leverages the zero-shot detection capabilities of foundation models to generate pseudo-labels for object instances across video sequences. By utilizing off-the-shelf VLMs, we automatically obtain bounding boxes for target objects with minimal human intervention. Human experts are involved only to refine these pseudo-labels when necessary, thereby reducing annotation costs compared to traditional methods. Building upon these pseudo-labels, we train a YOLOv8 segmentation network using a combination of the refined labels and publicly available datasets, as detailed in Section \ref{sec:method}. To enhance detection accuracy, especially in challenging conditions like high illumination, we incorporate pre-processing techniques such as relighting adjustments. During post-inference, depth-based filtering is applied to further refine the results. For object tracking, we employ a hybrid approach that integrates the Matching by Segment Anything (MASA) \cite{masa} adapter with the BoT-SORT \cite{aharon2022bot} algorithm.

Our experimental results demonstrate the effectiveness of the proposed methodology, achieving a \textbf{HOTA} score of 80.4\%, \textbf{MOTA} of 66.1\%, \textbf{Recall} of 74.0\%, and \textbf{Precision} of 90.7\%. These metrics indicate that our approach can successfully track sweet peppers without the need for extensive human intervention, highlighting the potential of leveraging foundation models for efficient and accurate object detection and tracking in agricultural settings. Further details of our overall framework and an analysis of key factors are explored in Section \ref{sec:method} and \ref{sec:experiments}.
\vspace{-1ex}
\section{Methodology}
\begin{figure}[t]
    \centering
    {{\includegraphics[width=1\textwidth]{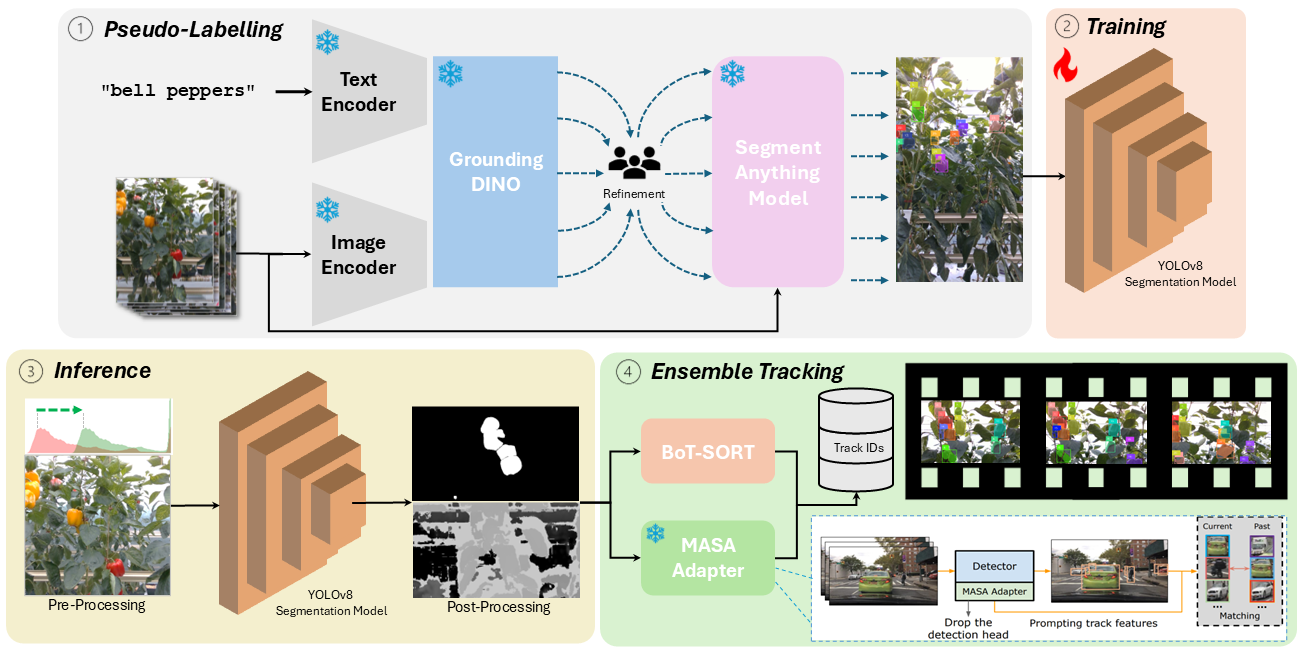}}}
    \caption{\textbf{The overall framework of TAP.} First, pseudo-labels are generated using a foundation model and refined by experts to ensure quality. SAM is then used to create segmentation masks based on these refined labels. The labels are subsequently employed to train a segmentation model for sweet pepper detection. Before inference, adaptive relighting compensates for illumination variance, while post-inference depth filtering separates object layers. Finally, a hybrid tracking system combines algorithms to assign unique tracking IDs and ensure consistency across frames.}
    \label{fig:flowchart}
    \vspace{-3ex}
\end{figure}

\vspace{-1ex}
\label{sec:method}
In this section, we present the methodology used to address the challenge of detecting and tracking small objects in a cluttered agricultural environment using weakly labeled data. Our approach consists of four main stages. \textcircled{1} \textbf{Weak Labels Acquisition} involves leveraging a foundation model to generate bounding boxes and segmentation masks, which are refined by human experts. Next, in \textcircled{2} \textbf{Mask \& Box Detection with YOLOv8}, we train a YOLOv8 segmentation network for object detection using publicly available datasets combined with our refined pseudo-labels. For \textcircled{3} \textbf{Pre- and Post-processing} step, we apply an \textit{adaptive relighting strategy} prior to inference to handle high illumination conditions, and perform \textit{depth filtering} after inference to separate foreground from background objects. \textcircled{4} \textbf{Hybrid Object Tracking with Ensemble Methods}, we employ an ensemble tracking method combining MASA adapter prompts and the BoT-SORT algorithm to associate unique tracking IDs for each detection across frames. The overall methodology is illustrated in Figure \ref{fig:flowchart}.

\subsubsection{Problem Formulation.}
Given a sequence of frames \( \mathcal{F} = \{f_1, f_2, \ldots, f_T\} \) from a video, the objective is to track small objects (e.g., sweet peppers) across consecutive frames. In each frame \( f_t \), potential object detections \( d_i^t \) are identified, where each detection \( d_i^t \) is characterized by a bounding box \( b(d_i^t) = (x_{\min}, y_{\min}, x_{\max}, y_{\max}) \), a detection probability \( p(d_i^t) \), and a segmentation mask \( m(d_i^t) \). The goal is to associate these detections across frames to form consistent object trajectories \( \mathcal{T}_k = \{d_k^1, d_k^2, \ldots, d_k^{N_k}\} \) for each object \( k \), where \( N_k \) denotes the number of frames in which object \( k \) appears. This tracking task must address challenges such as occlusion, clutter, and varying lighting conditions to maintain accurate and continuous trajectories throughout the video sequence.

\subsubsection{Weak Labels Acquisition with Foundation Models.}
\label{subsec:pseudo-labels}
We utilize Grounding DINO, a vision-language model, to perform zero-shot object detection using textual queries like \texttt{"bell\_pepper"}, enabling us to detect objects without prior training on our dataset. Unlike conventional detection models, Grounding DINO uses this semantic information to identify target objects. For each image sequence \( \mathcal{F} = \{f_1, f_2, \ldots, f_T\} \), where \( f_t \) represents the frame at timestamp \( t \), we define a sub-sampling interval \( I \) to extract \( N = \left\lfloor \frac{T}{I} \right\rfloor \) images, minimizing redundancy from visually similar frames. These sub-sampled images are input to Grounding DINO, which generates a set of pseudo-bounding boxes \( \mathcal{P} = \{p_1, p_2, \ldots, p_N\} \), with each \( p_i \) representing the coordinates of a detected object. To ensure quality, we consider only pseudo-labels with high confidence scores. The confidence score \( c_i \) for each detection \( p_i \) is computed as \( c_i = \sigma(l_i) \), where \( \sigma(\cdot) \) is the sigmoid function and \( l_i \) is the logit output associated with \( p_i \). We select detections satisfying \( c_i \geq \tau \), where \( \tau \) is a predefined threshold, resulting in a refined set \( \mathcal{P}' = \{p_i \mid c_i \geq \tau\} \). Human experts then further refine these selected pseudo-labels to enhance their accuracy, particularly in the cluttered environment. Each image is manually inspected, and only those with misaligned boxes or significant detection errors are refined, typically requiring less than 2 minutes per image. The refined bounding boxes serve as prompts for the Segment Anything Model (SAM), which generates precise segmentation masks \( \mathcal{M} = \{m_1, m_2, \ldots, m_N\} \) corresponding to the detected objects. These refined labels and masks are subsequently used in later stages for model training and detection.

\subsubsection{Mask \& Box Detection with YOLOv8.}
We train a YOLOv8 \cite{redmon2016you} model using the refined bounding boxes \( \mathcal{P}' \) and segmentation masks \( \mathcal{M} \) as inputs. While vision-language models like Grounding DINO exhibit strong zero-shot capabilities, we observed that YOLOv8 achieves superior performance due to its ability to generalize effectively with limited training data. This performance gain is attributed to YOLOv8's robustness against overfitting and reduced susceptibility to over-parameterization, both common challenges when fine-tuning large foundational models on small datasets. The YOLOv8 model is trained using a combination of three loss functions: classification loss \( \mathcal{L}_{\text{cls}} \), localization loss \( \mathcal{L}_{\text{loc}} \), and segmentation loss \( \mathcal{L}_{\text{seg}} \). The overall training objective is to minimize the total loss \( \mathcal{L}_{\text{overall}} \), defined as:
\begin{align}
    \mathcal{L}_{\text{overall}} = \lambda_{\text{cls}} \mathcal{L}_{\text{cls}} + \lambda_{\text{loc}} \mathcal{L}_{\text{loc}} + \lambda_{\text{seg}} \mathcal{L}_{\text{seg}},
\end{align}
where \( \lambda_{\text{cls}} \), \( \lambda_{\text{loc}} \), and \( \lambda_{\text{seg}} \) are hyperparameters that weight the contribution of each loss component.

\subsubsection{Adaptive Relighting Scheme.}
To enhance detection quality under varying lighting conditions, we implement an adaptive contrast and luminance adjustment scheme prior to inference. Our approach dynamically adjusts the luminance and contrast of each frame \( f_t \) based on its overall luminance \( \mathcal{L}(f_t) \). The adjusted luminance \( \mathcal{L}'(f_t) \) and contrast \( \mathcal{C}'(f_t) \) are computed as:
\begin{align}
    \mathcal{L}'(f_t) &= \alpha(\mathcal{L}(f_t)) \cdot \mathcal{L}(f_t), \\
    \mathcal{C}'(f_t) &= \beta(\mathcal{L}(f_t)) \cdot \mathcal{C}(f_t),
\end{align}
where \( \mathcal{C}(f_t) \) represents the original image contrast. The scaling factors \( \alpha(\mathcal{L}(f_t)) \) and \( \beta(\mathcal{L}(f_t)) \) are functions of the frame's luminance, designed to reduce luminance when it's high and boost contrast when it's low. By dynamically adjusting these factors, we enhance detection performance across a range of illumination levels by suppressing overexposed or underexposed areas and emphasizing object boundaries.

\subsubsection{Detection and Depth-Based Filtering.}
After training, we use the YOLOv8 model to detect sweet peppers in each video frame \( f_t \). Once detections are obtained, we apply the segmentation masks \( \mathcal{M} \), generated by YOLOv8, along with depth maps to further filter the detected objects. Let \( d_i \) denote the depth of instance \( i \), and let \( \tau_d \) be a threshold distinguishing foreground from background objects. Each object is classified as:
\begin{align}
    \text{Object}_i &=
    \begin{cases} 
    \mathsf{Foreground}, & \text{if } d_i < \tau_d \\
    \mathsf{Background}, & \text{if } d_i \geq \tau_d
    \end{cases}
\end{align}
For instances classified as foreground (\( d_i < \tau_d \)), we retain the predictions \( \hat{y}_i = (\hat{b}_i, \hat{p}_i, \hat{m}_i) \), where \( \hat{b}_i \), \( \hat{p}_i \), and \( \hat{m}_i \) are the predicted bounding box, detection probability, and segmentation mask, respectively. Instances classified as background (\( d_i \geq \tau_d \)) are discarded (\( \hat{y}_i = \varnothing \)).

\subsubsection{Hybrid Object Tracking with Ensemble Methods.}
After detecting the bounding boxes in each frame of the video sequence, we perform object tracking using a hybrid ensemble approach. The first component employs the BoT-SORT \cite{aharon2022bot} algorithm, a multi-object tracking method that associates detections across frames based on bounding box overlap and appearance features. Simultaneously, we utilize the detected bounding boxes as prompts for the Matching Anything by Segmenting Anything (MASA) \cite{masa} adapter, built on SAM. MASA enables instance-level correspondence tracking without requiring explicit video annotations by learning instance representations from dense, unlabeled data. This improves the model’s ability to track objects across frames by leveraging prompt-based supervision.
\vspace{-2ex}
\section{Experiments}
\label{sec:experiments}
\noindent\textbf{Dataset and Evaluation Metrics.} In this study, we utilize a combination of pseudo-labeled data and publicly available datasets to construct a comprehensive training set. On top of the pseudo-labels generated and refined described in Section \ref{subsec:pseudo-labels}, we also integrate additional datasets, including BUP20 \cite{smitt2024pagnerf} and a publicly available dataset from Kaggle \cite{montoya2022kaggle} to ensure diversity in object appearances and environments. For evaluation, we adopt the widely-used HOTA (Higher Order Tracking Accuracy) metric to assess tracking performance. HOTA evaluates both the detection and association performance of tracking algorithms. In addition to HOTA, we report results for MOTA (Multiple Object Tracking Accuracy) \cite{bernardin2008evaluating}, Recall, and Precision to provide a well-rounded assessment of our method's effectiveness.

\noindent\textbf{Implementation Details.} For the depth filtering step, the threshold \( \tau_d \) is set to 1200, ensuring that objects with a depth value \( d_i < \tau_d \) are retained as foreground, while the rest are discarded as background. For sub-sampling the video sequences, we use an interval \( I=5 \), meaning we select one frame out of every five to minimize overlap and redundancy in the training set, resulting in a total of 380 images. For training the YOLOv8 model, we use the default parameters provided by the Ultralytics project, with the following modifications: a batch size of 8 and all input images resized to 736x416 pixels.

\begin{table}[t]
\centering
\caption{\textbf{Tracking Performance}. We report the average Identification F1-score (IDF1), Precision (IDP), and Recall (IDR), as well as HOTA, MOTA, Recall (Rcll), and Precision (Prec) across all sequences in the evaluation set. The $\uparrow$ and $\downarrow$ indicate that lower or higher values mean better performance.}\vspace{-2ex}
\resizebox{0.9\linewidth}{!}{%
\begin{tabular}{c c c c c c c c}
    \toprule
    \multirow{2}{*}{\textbf{Method}} & \textbf{IDF1}[\%] & \textbf{IDP}[\%] & \textbf{IDR}[\%] & \textbf{HOTA}[\%] & \textbf{MOTA}[\%] & \textbf{Rcll}[\%] & \textbf{Prec}[\%] \\
    & $\uparrow$ avg & $\uparrow$ avg & $\uparrow$ avg & $\uparrow$ avg & $\uparrow$ avg & $\uparrow$ avg & $\uparrow$ avg \\
    \midrule
    \textbf{Ours} & 80.5 & 89.5 & 73.1 & 80.4 & 66.1 & 74.0 & 90.7 \\
    \bottomrule
\end{tabular}
}
\label{table:tracking_performance}
\end{table}

\noindent\textbf{Main Results.} Our model achieves a HOTA score of 80.4\%, MOTA of 66.1\%, Recall of 74.0\%, and Precision of 90.7\% as shown in Table \ref{table:tracking_performance}. The high \textbf{Precision} demonstrates effective reduction of false positives, primarily due to our depth filtering step, which eliminates background objects using an empirically selected depth threshold. We argue that pre-defining these settings would have allowed for further optimization, potentially improving performance. Contrary to this, the slightly lower \textbf{Recall} suggests that some true positives were missed, likely due to flickering detections in heavily occluded scenes. This may be attributed to MASA's \cite{masa} pre-training on static and unlabeled images, which limits its ability to perform well under scenarios with heavy occlusions.

\noindent\textbf{Ablation Study.}
We evaluate the effect of incorporating depth-based filtering into the detection pipeline by comparing results with and without depth filtering. The results, as shown in Table \ref{table:depth_filtering}, demonstrate that depth filtering significantly improves the quality of the final predictions by eliminating background objects and enhancing tracking performance in cluttered environments by separating overlapping objects.

\begin{table}[t]
\centering
\caption{\textbf{Ablation Study}. We report the impact of applying depth filtering on tracking performance.}\vspace{-2ex}
\resizebox{0.9\linewidth}{!}{%
\begin{tabular}{c c c c c c c c}
    \toprule
    \multirow{2}{*}{\textbf{Method}} & \textbf{IDF1}[\%] & \textbf{IDP}[\%] & \textbf{IDR}[\%] & \textbf{HOTA}[\%] & \textbf{MOTA}[\%] & \textbf{Rcll}[\%] & \textbf{Prec}[\%] \\
    & $\uparrow$ avg & $\uparrow$ avg & $\uparrow$ avg & $\uparrow$ avg & $\uparrow$ avg & $\uparrow$ avg & $\uparrow$ avg \\
    \midrule
    w/o Filtering & 58.6 & 62.2 & 55.4 & 63.4 & 22.4 & 55.8 & 62.7 \\
    \rowcolor{LightCyan!40} w/ Filtering & \textbf{80.5} & \textbf{89.5} & \textbf{73.1} & \textbf{80.4} & \textbf{66.1} & \textbf{74.0} & \textbf{90.7} \\
    \bottomrule
\end{tabular}
}
\label{table:depth_filtering}
\end{table}

\section{Conclusion}
In this study, we presented a methodology for detecting and tracking sweet peppers in agricultural environments by leveraging the zero-shot detection capabilities of large foundation models, thus minimizing the need for extensive manual annotation. By generating pseudo-labels with off-the-shelf vision-language models and involving human experts only for necessary refinements, we trained a YOLOv8 segmentation network using these labels alongside publicly available datasets. Employing pre-processing techniques like relighting adjustments and followed by an ensemble tracking algorithm, we effectively assigned unique IDs across video frames. We hope that our work provides valuable insights for future research in agricultural computer vision.


%
%
\bibliographystyle{splncs04}
\bibliography{main}
\end{document}